\documentclass[11pt,a4paper]{article}

\usepackage{fontspec}
\usepackage{polyglossia}
\setmainlanguage{english}

\newfontfamily\devfont{NotoSansDevanagari-Regular.ttf}[
  Path = ./fonts/ ,
  BoldFont = NotoSansDevanagari-Bold.ttf ,
  Script = Devanagari ]
\newcommand{\dev}[1]{{\devfont #1}}

\usepackage{geometry}
\usepackage{microtype}
\usepackage{amssymb}
\usepackage[round,authoryear]{natbib}
\bibpunct{(}{)}{;}{a}{}{,}
\usepackage{booktabs}
\usepackage{array}
\usepackage{longtable}
\usepackage{enumitem}
\usepackage{xcolor}
\usepackage{graphicx}
\usepackage{tikz}
\usetikzlibrary{arrows.meta,positioning,fit,calc}
\definecolor{stagefill}{RGB}{237,241,246}
\definecolor{stageline}{RGB}{74,92,120}
\definecolor{iofill}{RGB}{231,239,235}
\definecolor{ioline}{RGB}{58,104,84}
\definecolor{resfill}{RGB}{247,243,236}
\definecolor{resline}{RGB}{146,122,84}
\definecolor{humanfill}{RGB}{250,240,222}
\definecolor{humanline}{RGB}{176,122,40}
\definecolor{arrowcol}{RGB}{60,72,92}
\usepackage{hyperref}
\usepackage{xurl} 
\hypersetup{colorlinks=true, linkcolor=black, citecolor=black,
            urlcolor={blue!55!black}, pdftitle={A Word-Level Digital
            Reader of the Prasthanatrayi with Sankara's Bhasya}}
\urldef{\liveurl}\url{https://cs.rkmvu.ac.in/~tamal/prasthanatrayi/prasthanatrayi-sankara-bhasya-detail.html}
\usepackage{abstract}
\usepackage{titlesec}
\titleformat{\section}{\large\bfseries}{\thesection}{0.6em}{}
\titleformat{\subsection}{\normalsize\bfseries}{\thesubsection}{0.6em}{}

\title{\bfseries A Word-Level Digital Reader of the Prasth\=anatray\={\i}
with \'Sa\.nkara's Bh\=a\d{s}ya:\\[2pt]
\large Corpus, Method, and an Open, Offline Reading Aid for the
Advaita Ved\=anta Canon}

\author{%
  Tamal Maharaj\thanks{Corresponding author: \texttt{tamal@gm.rkmvu.ac.in}.
    A live instance of the reader is publicly available at \liveurl}\\
  \small Department of Computer Science\\
  \small Ramakrishna Mission Vivekananda Educational and Research Institute\\
  \small Belur Math, Howrah, West Bengal, India
}
\date{July 2026}

\newcommand{\iast}[1]{\textit{#1}}
\newcommand{\Sankara}{\'Sa\.nkara}
\newcommand{\bhasya}{bh\=a\d{s}ya}
\newcommand{\Prasthana}{Prasth\=anatray\={\i}}
\newcommand{\Upanisad}{Upani\d{s}ad}
\newcommand{\Gita}{Bhagavadg\={\i}t\=a}
\newcommand{\Brahmasutra}{Brahmas\=utra}
\newcommand{\padaccheda}{padaccheda}
\newcommand{\viccheda}{sandhi-viccheda}

\begin{document}
\maketitle

\begin{abstract}
\noindent
The \Prasthana{}---the ten principal \Upanisad{}s, the \Brahmasutra{}, and the
\Gita{}, together with \Sankara{}'s commentaries (\bhasya{})---is the
foundational corpus of Advaita Ved\=anta. For the non-specialist, and even
for the student, the text is difficult to read at the word level: continuous
euphonic combination (\iast{sandhi}), long nominal compounds
(\iast{sam\=asa}), and dense scholastic prose obscure where one word ends and
the next begins, and what each word means grammatically. We present an open,
fully offline, word-level digital reader of the entire \Prasthana{} with
\Sankara{}'s \bhasya{}. In the reader, \emph{every} word---both of the root
text (\iast{m\=ula}) and of the commentary---is clickable and resolves to a
pop-up giving its euphonic split (\padaccheda{}), morphological and syntactic
analysis, and gloss. Because every word carries a lemma, the reader doubles as a
\emph{concordance}: a search on a dictionary headword retrieves all of that
word's inflected and \iast{sandhi}-hidden occurrences---and its occurrences
inside compounds---across both layers. The resource covers thirteen commentarial units
(2{,}971 verses, s\=utras, and prose sections; 36{,}881 analysed
word-occurrences of root text) and a global dictionary of 95{,}587 distinct
commentarial surface forms. We describe the corpus, the hybrid analysis
pipeline---a rule-based \viccheda{} engine layered over an authoritative
inflected-form lexicon and attested-corpus look-ups, with large-language-model
(LLM)-assisted word analysis subjected to an adversarial two-pass verification
protocol---and a durable human-expert review loop that lets a Sanskritist
correct machine output once and have the correction survive all subsequent
regenerations. An intrinsic evaluation against independent Sanskrit resources
finds that high-confidence analyses agree with an authoritative inflectional
lexicon on over 99\,\% of attested forms, and an independent, band-blind
adjudication confirms that analysis quality degrades predictably across the
confidence bands, with errors concentrated in the low-confidence tier that the
review loop targets. The reader is a single, self-contained HTML file requiring
no server, no installation, and no network access, and is offered as a freely
redistributable teaching and reading aid.

\vspace{0.6em}
\noindent\textbf{Keywords:} Advaita Ved\=anta; \Sankara{}; \Prasthana{};
\Upanisad{}; digital humanities; computational Sanskrit; \viccheda{};
morphological analysis; digital philology; open educational resources.
\end{abstract}

\section{Introduction}

The three \emph{prasth\=ana}s or ``points of departure'' of Ved\=anta---the
\iast{\'sruti-prasth\=ana} (the principal \Upanisad{}s), the
\iast{ny\=aya-prasth\=ana} (the \Brahmasutra{} of B\=adar\=aya\d{n}a), and the
\iast{sm\d{r}ti-prasth\=ana} (the \Gita{})---constitute the canon on which
every classical Ved\=antic school grounds its authority. \Sankara{}'s
commentaries (\bhasya{}) on these three are, for the Advaita tradition, the
paradigmatic exposition; they are read, memorised, and debated to this day.

Yet the corpus is hard to \emph{read} at the granularity a learner needs.
Classical Sanskrit is written in continuous euphonic combination: adjacent
words fuse at their boundaries by regular phonological rules (\iast{sandhi}),
and nominal compounds (\iast{sam\=asa}) can run to many members without
internal spacing. A single orthographic ``word'' on the page,
e.g.\ \dev{धूमेनाव्रियते}, may in fact be two or three words
(\iast{dh\=umena \=avriyate}), and \Sankara{}'s prose---syntactically
intricate, allusive, and studded with unmarked scripture citations---raises
this difficulty to a further power. The reader who cannot perform
\viccheda{} (the ``un-joining'' of \iast{sandhi}) and identify each word's
stem, inflection, and syntactic role cannot in practice consult a dictionary,
because the surface form does not appear there.

Traditionally this competence is transmitted orally, over years, from teacher
to student. Printed word-by-word editions (\iast{anvaya} and \iast{\d{t}\={\i}k\=a})
exist for a few celebrated texts---above all the \Gita{}---but no uniform,
word-level apparatus exists across the whole \Prasthana{}, and none exists at
all for the running prose of \Sankara{}'s commentary, which is many times the
length of the root text it explains.

This paper reports a resource that closes that gap. We have built an open,
offline, word-level digital reader in which every word of both the root text
and the commentary is directly interrogable. Clicking a word opens an analysis
pop-up: the euphonic split, the constituent stems, the grammatical description
(part of speech, gender/number/case for nominals; class, tense/mood, person,
number for verbs; the syntactic \iast{k\=araka} role where relevant), and an
English gloss. Crucially, because the written form of a Sanskrit word rarely
matches its dictionary headword---\iast{sandhi}, inflection, and compounding all
disguise it---the reader also turns the edition into a \emph{lemma-indexed
concordance}: a search on a stem retrieves every one of its inflected and
\iast{sandhi}-hidden occurrences, and (through a complementary \emph{split-aware}
search) its occurrences buried inside compounds, across both the root text and
the whole of the commentary. Whole-word highlighting spans both textual layers.

Our contribution is fourfold:
\begin{enumerate}[leftmargin=1.4em,itemsep=0.15em]
  \item \textbf{A corpus resource:} a uniformly word-analysed digital edition of the
        complete \Prasthana{} \emph{with} the running \Sankara{}-\bhasya{}, at a
        scale (95{,}587 distinct commentarial forms) not previously available in
        interrogable form.
  \item \textbf{A discovery tool:} because every word is lemma-tagged, the edition
        is searchable as a \emph{concordance by dictionary headword}---retrieving a
        term's inflected, \iast{sandhi}-hidden, and compound-internal occurrences
        across both layers---a form of retrieval that plain full-text search over
        Devan\=agar\={\i} cannot provide.
  \item \textbf{A method:} a reproducible, hybrid pipeline that combines
        deterministic linguistic resources (a rule-based \viccheda{} engine, a
        927{,}000-form inflectional lexicon, attested-corpus look-ups) with
        LLM-assisted analysis under an adversarial two-pass verification
        protocol, and a durable human-in-the-loop correction overlay.
  \item \textbf{An artefact:} a single self-contained HTML file, dependency-free
        and freely redistributable, designed as a teaching and self-study aid
        and as a platform for scholarly correction.
\end{enumerate}

We stress at the outset that the machine analysis is a \emph{scaffold}, not an
oracle. Final scholarly authority rests with the human editor; the system's
role is to produce a consistent first-pass analysis over an otherwise
prohibitive volume of text, to flag its own least-confident output, and to make
expert corrections cheap, durable, and cumulative.

\section{Background and Related Work}

\paragraph{Computational Sanskrit.}
The last two decades have produced mature tools for Sanskrit segmentation and
morphology. Huet's \emph{Sanskrit Heritage} engine and its reader provide a
finite-state morphological analyser and a segmenter grounded in P\=a\d{n}inian
description \citep{huet2005,goyal2016}. The \emph{Digital Corpus of Sanskrit}
(DCS) offers a large, morphologically and lexically annotated
corpus~\citep{hellwig2010dcs}, and recent neural systems have advanced
data-driven word segmentation and dependency parsing~\citep{hellwig2018,krishna2018,sandhan2021}.
The biennial \emph{Sanskrit Computational Linguistics} symposia collect much of
this work~\citep{scl2009}. Our system is not a competitor to these analysers;
it \emph{consumes} them---most directly, an inflected-form lexicon extracted
from the Heritage engine and \padaccheda{} attestations drawn from annotated
corpora---and adds an LLM-assisted analysis and verification layer on top,
oriented toward a specific, closed, high-value corpus.

\paragraph{Digital editions and lexica.}
Digital text archives such as GRETIL and the Muktabodha Indological Research
Institute's digital library make the \Prasthana{} available as searchable text,
and the Cologne Digital Sanskrit Dictionaries project has made the standard
lexica (Monier-Williams, Apte, and others) machine-readable~\citep{cologne}.
For the \Gita{} specifically, the IIT Kanpur ``Gita Supersite'' presents
multiple commentaries side by side. These resources supply
\emph{text} and \emph{glosses} but not, in general, a uniform
per-word grammatical apparatus over running commentarial prose; that apparatus
is what the present work adds, and it draws its dictionary glosses from the
Cologne lexica.

\paragraph{LLMs for classical philology.}
Large language models have recently been applied to low-resource and classical
languages for translation, restoration, and annotation. Their promise for
Sanskrit is a fluent first-pass analysis at scale; their danger is confident
error. Our design responds to exactly this trade-off by (i) constraining and
cross-checking LLM output against deterministic resources, (ii) running an
adversarial ``refute'' pass whose sole job is to find mistakes in the first
pass, (iii) attaching an explicit confidence label to every analysis, and
(iv) routing low-confidence output to a human expert whose verdicts are stored
permanently. The methodological contribution of this paper is as much this
\emph{verification harness} as the corpus itself.

\section{The Corpus}

The reader covers the complete \Prasthana{} as commented by \Sankara{}:
the ten principal \Upanisad{}s (with \Sankara{}'s two separate commentaries on
the Kena, the \iast{padabh\=a\d{s}ya} and the \iast{v\=akyabh\=a\d{s}ya}), the
\Brahmasutra{}, and the \Gita{}---thirteen commentarial units in all.
The Māṇḍūkya unit additionally carries Gau\d{d}ap\=ada's
\iast{K\=arik\=a}, which \Sankara{} also comments upon. Table~\ref{tab:corpus}
gives the breakdown.

\begin{table}[htbp]
\centering
\small
\begin{tabular}{@{}l l r@{}}
\toprule
\textbf{Text} & \textbf{Class} & \textbf{Verses / s\=utras / sections} \\
\midrule
\={I}\'s\=a \Upanisad{}                         & \'sruti & 18 \\
Kena \Upanisad{} (\iast{padabh\=a\d{s}ya})      & \'sruti & 35 \\
Kena \Upanisad{} (\iast{v\=akyabh\=a\d{s}ya})   & \'sruti & 35 \\
Ka\d{t}ha \Upanisad{}                          & \'sruti & 120 \\
Pra\'sna \Upanisad{}                           & \'sruti & 67 \\
Mu\d{n}\d{d}aka \Upanisad{}                    & \'sruti & 65 \\
M\=a\d{n}\d{d}\=ukya \Upanisad{} (+ K\=arik\=a) & \'sruti & 227 \\
Aitareya \Upanisad{}                           & \'sruti & 33 \\
Taittir\={\i}ya \Upanisad{}                    & \'sruti & 53 \\
Ch\=andogya \Upanisad{}                        & \'sruti & 629 \\
B\d{r}had\=ara\d{n}yaka \Upanisad{}            & \'sruti & 434 \\
\Brahmasutra{}                                 & ny\=aya & 555 \\
\Gita{}                                        & sm\d{r}ti & 700 \\
\midrule
\textbf{Total}                                 &         & \textbf{2{,}971} \\
\bottomrule
\end{tabular}
\caption{The corpus. ``Verses/s\=utras/sections'' counts addressable
root-text units; prose \Upanisad{}s (Ch\=andogya, B\d{r}had\=ara\d{n}yaka)
are counted by section. The M\=a\d{n}\d{d}\=ukya total includes
Gau\d{d}ap\=ada's \iast{K\=arik\=a}.}
\label{tab:corpus}
\end{table}

The single source of truth is a corpus file that stores, for each addressable
unit, the root-text words, the running \bhasya{} prose, and the
\iast{sambandha-bh\=a\d{s}ya} (the introductory ``connecting'' passages) as
structured data. All derived artefacts---the analysed word data and the reader
HTML---are \emph{built} from this source by scripts; the built files are never
hand-edited, which keeps the pipeline reproducible and the source authoritative.

\section{System Overview}

The deliverable is a pair of self-contained HTML files that share a common
build. The \emph{m\=ula reader} makes every root-text word clickable. The
\emph{detail reader} is a superset: it additionally makes every word of
\Sankara{}'s commentary clickable. Both are single files that open directly in
any modern browser with no server and no network connection; the detail reader
embeds its 95{,}587-entry commentary dictionary inline as JSON.

Two design decisions are worth naming because they shape the whole system.

\paragraph{Caret-based word resolution, not per-word markup.}
A naive way to make words clickable is to wrap each in its own HTML element.
For a corpus this size that would bloat the file, and---more importantly---it
would fragment the text nodes and break running-text search and highlighting.
Instead, a click anywhere in the prose is resolved to the word under the cursor
at read time, by taking the maximal run of Devan\=agar\={\i} characters around
the caret position. The tokeniser that does this is character-for-character
identical to the one used offline during extraction, so a clicked surface is
guaranteed to match a dictionary key. The DOM is left pristine, and phrase
search and highlighting continue to work over unbroken text.

\paragraph{``Route B'': analyse each distinct surface once.}
\Sankara{}'s prose is highly repetitive at the word level: the same inflected
form (\iast{\=atman}, \iast{brahma\d{n}a\d{h}}, \iast{tasm\=at}, \dots) recurs
thousands of times. Rather than analyse every \emph{occurrence}, we build a
\emph{global dictionary of distinct surface forms} and analyse each exactly
once. This reduces the analysis burden by more than an order of magnitude and
guarantees internal consistency (the same surface always receives the same
analysis), at the cost of not disambiguating a form by its local context---a
trade-off we return to in \S\ref{sec:limitations}. The root text, being far
smaller and where context matters more, is instead analysed
\emph{per-occurrence} and indexed by position.

\section{Method: the Analysis Pipeline}

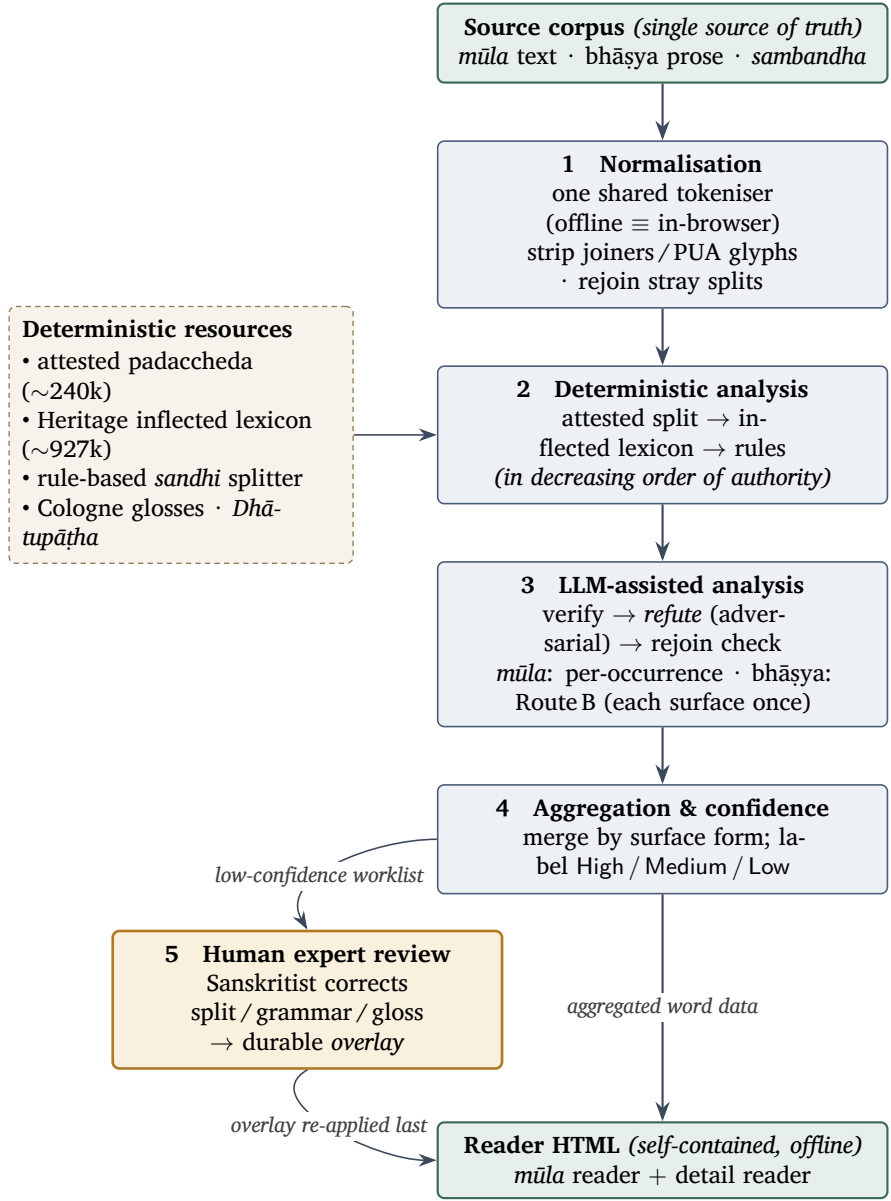
\begin{figure}[tbp]
\centering
\footnotesize
\begin{tikzpicture}[
  node distance = 7.5mm and 11mm,
  every node/.style = {font=\footnotesize},
  stage/.style = {rectangle, rounded corners=2.5pt, draw=stageline, line width=0.6pt,
                   fill=stagefill, align=center, text width=56mm, inner sep=5pt},
  io/.style    = {rectangle, rounded corners=2.5pt, draw=ioline, line width=0.8pt,
                   fill=iofill, align=center, text width=56mm, inner sep=5pt},
  res/.style   = {rectangle, rounded corners=2.5pt, draw=resline, line width=0.5pt,
                   dash pattern=on 2pt off 1.6pt, fill=resfill, align=left,
                   text width=42mm, inner sep=5pt},
  human/.style = {rectangle, rounded corners=2.5pt, draw=humanline, line width=1pt,
                   fill=humanfill, align=center, text width=48mm, inner sep=5pt},
  flow/.style  = {-{Stealth[length=2.6mm,width=1.9mm]}, draw=arrowcol, line width=0.8pt},
  side/.style  = {-{Stealth[length=2.4mm,width=1.7mm]}, draw=arrowcol, line width=0.6pt},
  lbl/.style   = {font=\scriptsize\itshape, inner sep=1.5pt, fill=white, text=black!75}
]
\node[io]    (corpus) {\textbf{Source corpus} \emph{(single source of truth)}\\
                       \iast{m\=ula} text \,\textperiodcentered\, \bhasya{} prose \,\textperiodcentered\, \iast{sambandha}};
\node[stage] (norm) [below=of corpus]
   {\textbf{1\quad Normalisation}\\ one shared tokeniser (offline $\equiv$ in-browser)\\
    strip joiners\,/\,PUA glyphs \,\textperiodcentered\, rejoin stray splits};
\node[stage] (det) [below=of norm]
   {\textbf{2\quad Deterministic analysis}\\ attested split $\to$ inflected lexicon $\to$ rules\\
    \emph{(in decreasing order of authority)}};
\node[stage] (llm) [below=of det]
   {\textbf{3\quad LLM-assisted analysis}\\ verify $\to$ \emph{refute} (adversarial) $\to$ rejoin check\\
    \iast{m\=ula}: per-occurrence \,\textperiodcentered\, \bhasya{}: Route\,B (each surface once)};
\node[stage] (agg) [below=of llm]
   {\textbf{4\quad Aggregation \& confidence}\\ merge by surface form;
    label \textsf{High\,/\,Medium\,/\,Low}};
\node[io]    (out) [below=30mm of agg]
   {\textbf{Reader HTML} \emph{(self-contained, offline)}\\
    \iast{m\=ula} reader $+$ detail reader};
\node[res] (resrc) [left=of det]
   {\textbf{Deterministic resources}\\[1pt]
    \textbullet\ attested \padaccheda{} ($\sim$240k)\\
    \textbullet\ Heritage inflected lexicon ($\sim$927k)\\
    \textbullet\ rule-based \iast{sandhi} splitter\\
    \textbullet\ Cologne glosses \,\textperiodcentered\, \iast{Dh\=atup\=a\d{t}ha}};
\node[human] (rev) at ($(agg)!0.5!(out) + (-47mm,0)$)
   {\textbf{5\quad Human expert review}\\ Sanskritist corrects split\,/\,grammar\,/\,gloss\\
    $\to$ durable \emph{overlay}};
\draw[flow] (corpus) -- (norm);
\draw[flow] (norm)   -- (det);
\draw[flow] (det)    -- (llm);
\draw[flow] (llm)    -- (agg);
\draw[flow] (agg)    -- (out) node[lbl,midway]{aggregated word data};
\draw[side] (resrc.east) -- (det.west);
\draw[side] (agg.west) to[out=180,in=120]
      node[lbl,pos=0.62]{low-confidence worklist} (rev.north);
\draw[side] (rev.south) to[out=-150,in=180]
      node[lbl,pos=0.5]{overlay re-applied last} (out.west);
\end{tikzpicture}
\caption{The analysis pipeline. A single source corpus is normalised, then passed
through a deterministic layer (attested splits, an inflected-form lexicon, and a
rule-based splitter, tried in that order) with an LLM-assisted layer on top under an
adversarial \emph{verify}\,--\,\emph{refute}\,--\,rejoin harness. Results are aggregated by
surface form and confidence-banded; the low-confidence tail is routed to a durable
human-review overlay that is re-applied on every rebuild, and the analysed data is
emitted as two self-contained reader files. Root text (\iast{m\=ula}) is analysed per
occurrence in verse context; commentary (\bhasya{}) surfaces are analysed once each
(``Route~B'').}
\label{fig:pipeline}
\end{figure}

The pipeline has five stages (Figure~\ref{fig:pipeline}): normalisation,
deterministic linguistic analysis, LLM-assisted analysis under adversarial
verification, aggregation, and human expert review. We describe each.

\subsection{Text normalisation}
Digitised Sanskrit sources carry rendering artefacts that a tokeniser must not
mistake for word boundaries: zero-width joiners and non-joiners inserted to
force explicit-halant conjunct rendering (e.g.\ in \dev{साङ्ख्य}); private-use
glyphs left by legacy fonts; and stray spaces that split a single word across
two tokens. A single shared normaliser is applied \emph{both} during offline
extraction \emph{and}, at build time, to the text embedded in the reader, so
that dictionary keys and in-browser tokens agree exactly. It strips joiners and
private-use glyphs, preserves the Vedic anusv\=ara \dev{ꣳ} (U+A8F3) as a
word character, and performs a lexicon-assisted rejoining of stray-split words
(e.g.\ \dev{यथै व} $\rightarrow$ \dev{यथैव}). The word-character class is
widened beyond the basic Devan\=agar\={\i} block to include the Vedic and
Devan\=agar\={\i}-Extended ranges, and the same class is used by the Python
extractor and by both in-browser tokenisers---this agreement is a hard
invariant of the system.

\subsection{Deterministic linguistic analysis}
Before any LLM is invoked, the system draws on three deterministic resources,
in order of authority:
\begin{enumerate}[leftmargin=1.4em,itemsep=0.15em]
  \item \textbf{Attested-corpus \padaccheda{}.} Splits attested in annotated
        corpora (some 240{,}000 word entries from a public Sanskrit corpus,
        supplemented by the Digital Corpus of Sanskrit for prose and
        \iast{dar\'sana} registers) are used first: an attested split is
        preferred to a computed one.
  \item \textbf{An authoritative inflected-form lexicon.} A lexicon of roughly
        927{,}000 inflected forms, extracted from Huet's Heritage engine,
        supplies real morphological readings (stem, gender, case/number, or
        verbal features) for recognised surfaces.
  \item \textbf{A rule-based \viccheda{} engine.} For forms not otherwise
        resolved, an offline, rule-based sandhi-splitter proposes boundaries.
        Evaluated in isolation against a \Gita{} ground truth it is a modest
        baseline; its role in the layered system is a fallback, not the primary
        analyser.
\end{enumerate}
Glosses are drawn from the standard lexica (Monier-Williams, Apte, and others)
in a rebuilt structured form, and bare verbal roots are glossed from the
\iast{Dh\=atup\=a\d{t}ha}.

\subsection{LLM-assisted analysis with adversarial verification}
\label{sec:llm}
The residue---forms not confidently resolved by the deterministic layer, and,
for the root text, every word in its verse context---is analysed with the help
of a large language model (in our runs, Anthropic's Claude Opus), which is well
suited to the joint task of splitting a compound, assigning grammar, and
glossing in one pass while ``seeing'' the surrounding text. Crucially, the LLM
is embedded in a verification harness rather than trusted directly.

\paragraph{Root text: two-pass verify-and-refute.}
Every root-text verse is processed by \emph{two} independent LLM passes. The
first (\emph{verify}) produces, for each word, a proposed split, grammar,
gloss, and a self-assessed confidence. The second (\emph{refute}) is
adversarial: a separate agent is shown the verse, the relevant \Sankara{}
commentary, and the first pass's analysis, and is instructed \emph{to find and
correct errors}---to try to refute each split. Only analyses that survive this
adversarial pass are retained. This two-pass protocol was run to completion
over all thirteen texts (36{,}881 word-occurrences); it surfaced and corrected
a class of plausible-but-wrong splits that a single pass accepts. Every
proposed split is additionally checked mechanically for the property that its
parts \emph{rejoin}, under forward-sandhi, to the original surface---a cheap
consistency filter against splits that do not add up.

\paragraph{Commentary: batched surface analysis.}
The 95{,}587 distinct commentary surfaces are analysed in batches, each batch
handled by one agent instructed to return, per surface, a structured record of
split, grammar, gloss, and confidence. Parenthetical scripture-references
embedded in the prose are stripped before extraction so that citation
abbreviations are not mistaken for words to analyse; citations remain live
navigational links in the reader.

\subsection{Aggregation and confidence}
The per-batch results are merged into a single dictionary keyed by surface
form, retaining only surfaces that occur in the source (so that stray citation
tokens are dropped). Each entry carries a confidence label. Table~\ref{tab:conf}
gives the distribution over the 95{,}587 commentary surfaces. The
low-confidence tail is not hidden: it is exported as a review worklist
(\S\ref{sec:review}).

\begin{table}[htbp]
\centering
\small
\begin{tabular}{@{}l r r@{}}
\toprule
\textbf{Confidence} & \textbf{Surfaces} & \textbf{Share} \\
\midrule
High (\iast{h})   & 79{,}001 & 82.6\,\% \\
Medium (\iast{m}) & 14{,}899 & 15.6\,\% \\
Low (\iast{l})    &  1{,}687 &  1.8\,\% \\
\midrule
\textbf{Total}    & \textbf{95{,}587} & 100\,\% \\
\bottomrule
\end{tabular}
\caption{Self-assessed confidence over the commentary surface dictionary. The
1{,}687 low-confidence forms constitute the standing expert-review worklist.}
\label{tab:conf}
\end{table}

\subsection{The human expert-review loop}
\label{sec:review}
Machine analysis at this scale will contain errors, and a resource meant for
scholarship must let a scholar fix them---once, and permanently. The system
therefore closes with a human-in-the-loop correction cycle designed so that
expert effort is never wasted:
\begin{enumerate}[leftmargin=1.4em,itemsep=0.15em]
  \item The low-confidence surfaces are exported, frequency-sorted, to a
        spreadsheet. Each row carries one to three real occurrences of the form
        \emph{in context} (the surrounding \bhasya{} text with a reference),
        the machine's proposed split/grammar/gloss, and blank columns for the
        reviewer's corrections and a verdict.
  \item A Sanskritist fills in a corrected split, grammar, or gloss (any
        subset), or simply marks a row ``ok''.
  \item The corrections are folded into a durable \emph{overlay} file, keyed by
        surface and accumulated across review rounds.
  \item On every subsequent rebuild, the overlay is applied \emph{last},
        overriding machine output and marking the entry ``reviewed''; the reader
        displays such entries with a distinct ``reviewed \checkmark'' badge.
\end{enumerate}
Because corrections live in a separate overlay and are re-applied after every
regeneration, they survive any future re-run of the machine pipeline. Expert
judgement is thus cumulative and monotone: the resource can only get more
accurate as scholars use it, and no correction is ever lost to a rebuild. The
context-extraction step that populates the worklist runs the \emph{same}
normalisation-and-tokenisation pipeline as the dictionary extraction, so that
every flagged form is guaranteed to be locatable in the corpus with its
surrounding text.

\section{The Reading Interface}

The reader is designed for the learner and the scholar alike. Its principal
features:
\begin{itemize}[leftmargin=1.4em,itemsep=0.15em]
  \item \textbf{Click-for-analysis.} A click on any word of the root text or
        commentary opens a pop-up with its split (\padaccheda{}), grammatical
        analysis, gloss, and confidence (with the ``reviewed'' badge where a
        scholar has confirmed it). Citations are not analysed but remain live
        cross-reference links.
  \item \textbf{Lemma-aware and split-aware search.} Beyond surface strings, a
        search can match a word's \emph{lemma} (dictionary headword), so a query
        on a stem returns all of its inflected forms; the \emph{constituents} of
        a split, so a query on a compound member finds the compounds that contain
        it even when \iast{sandhi} hides the boundary; and suppletive pronominal
        stems are canonicalised, so a search on one stem gathers all its oblique
        forms. Search can be scoped to the root text, the commentary, or both.
  \item \textbf{Whole-word highlighting} across both textual layers, with
        independent toggles for the root text and the commentary, so that a
        matched word is highlighted in full even where a sandhi boundary falls
        inside it.
  \item \textbf{Scoped search} over any chosen subset of the thirteen texts,
        including the introductory \iast{sambandha-bh\=a\d{s}ya} passages.
  \item \textbf{Verse-line formatting} for the metrical texts, which break each
        \iast{\'sloka} at the \iast{da\d{n}\d{d}a} into half-verse lines for
        readability, while prose texts are left as running text.
\end{itemize}

\smallskip\noindent\emph{Why lemma search matters.} In Sanskrit the form on the
page almost never matches the form in a dictionary: \iast{sandhi} fuses word
boundaries, inflection rewrites endings, and compounding buries a word inside a
longer one, so a reader who types a dictionary form into an ordinary text search
finds almost nothing. Because this edition stores a lemma for every one of its
words, a single query on the headword \dev{आत्मन्} (\iast{\=atman}, ``self'')
instead surfaces \dev{आत्मा}, \dev{आत्मानम्}, \dev{आत्मनः}, \dev{आत्मनि}, \dots\
in every case and number, and---through the split index---its occurrences
\emph{inside} compounds such as \dev{परमात्मा} (\iast{parama-\=atman}) and
\dev{प्रत्यगात्मन्} (\iast{pratyag-\=atman}), throughout both the root text and
the entirety of \Sankara{}'s commentary. The reader is thus not merely a text to
read but a \emph{concordance} of the \Prasthana{}: a scholar can trace a term or
concept through the whole corpus by its dictionary form---a retrieval that plain
full-text search over the surface text cannot perform.

The interface language is English throughout, with Devan\=agar\={\i} reserved
for the Sanskrit text and grammatical labels, so that the reader is usable by
students without prior exposure to a modern Indian language.

\section{Evaluation}
\label{sec:eval}

We evaluate the analyses two ways, both stratified by the self-assessed
confidence band: (i) a \emph{deterministic} agreement study against independent
linguistic resources, run over the entire analysed corpus, and (ii) an
\emph{independent adjudication} of a blind stratified sample. The two are
complementary: the first is non-circular but can only score the subset of forms
covered by the reference resources; the second covers every dimension including
the gloss but is itself a language model and so a weaker, secondary signal. The
whole study is reproducible from a single script.\footnote{\texttt{tools/eval\_accuracy.py};
the adjudication sample and verdicts are archived alongside it.}

\subsection{Deterministic agreement with independent resources}
For every analysis we compute three checks against resources that played no part
in producing it:
\begin{itemize}[leftmargin=1.4em,itemsep=0.1em]
  \item \textbf{Morphology} --- for single-pada nominals, whether the analysis's
        (case, number, gender) matches a reading attested for the word in the
        Heritage inflected-form lexicon. Ambiguous analyses (e.g.\ nominative/
        accusative for a neuter) are credited if \emph{any} alternative matches;
        the lexicon is queried on both the surface token and its sandhi-restored
        form, allowing for the underlying \iast{-s}/\iast{-r}/\iast{-m} of a
        final visarga or anusv\=ara.
  \item \textbf{Verbal root} --- for finite verbs, whether the analysed root
        matches a Heritage verbal reading of the form.
  \item \textbf{Segmentation} --- for multi-pada surfaces attested in an
        independent annotated corpus, agreement between the analysis's split and
        the attested split (exact lemma-set match and mean lemma Jaccard).
\end{itemize}
Table~\ref{tab:det} reports the results on the bh\=a\d{s}ya dictionary. Two
findings stand out. First, agreement is high where an independent authority
exists: high-confidence morphology matches Heritage on 99.6\,\% of attested
forms by (case, number) and 99.4\,\% including gender; a manual audit of the
residual $<$1\,\% found that most are not analysis errors at all but gaps in the
lexicon (e.g.\ \iast{maghavata\d{h}} ``of Indra,'' correctly analysed as
masculine but listed only as neuter in the lexicon), so the figure is if
anything a slight under-estimate. Second, the confidence bands are
\emph{meaningful}: agreement falls monotonically from the high to the low band
on every measure, confirming that the self-assessed labels rank quality and that
the review worklist is correctly targeted at the least reliable analyses. The
mūla layer, analysed per-occurrence under the two-pass protocol, agrees with
Heritage on 99.2\,\% of attested nominals (case+number+gender) and 90.0\,\% of
verbal roots---an independent confirmation of the verified root-text data.

\begin{table}[htbp]
\centering
\small
\begin{tabular}{@{}l r r r r r@{}}
\toprule
 & \multicolumn{2}{c}{\textbf{Morphology (Heritage)}} & \textbf{Verb root} & \multicolumn{2}{c}{\textbf{Split (corpus)}}\\
\cmidrule(lr){2-3}\cmidrule(lr){4-4}\cmidrule(lr){5-6}
\textbf{Band} & \textbf{attested} & \textbf{c+n+g \%} & \textbf{\% (n)} & \textbf{set \% (n)} & \textbf{Jaccard}\\
\midrule
High   & 9{,}353 & 99.4 & 89.9 (1{,}296) & 74.8 (1{,}669) & 0.82\\
Medium & 1{,}438 & 95.5 & 87.4 (350)     & 58.7 (126)     & 0.70\\
Low    &   126   & 88.1 & 75.9 (54)      & 75.0 (16)      & 0.75\\
\midrule
All    & 10{,}917 & 98.8 & 88.9 (1{,}700) & 73.7 (1{,}811) & 0.81\\
\bottomrule
\end{tabular}
\caption{Deterministic agreement of the bh\=a\d{s}ya analyses with independent
resources, by confidence band. ``attested'' = single-pada nominals found in the
Heritage lexicon (of 40{,}417 applicable; the remainder are compounds/derivatives
absent from the lexicon as whole forms). ``Split (corpus)'' is measured on the
multi-pada surfaces attested in an independent corpus. Small $n$ in the low band
makes its split figures noisy.}
\label{tab:det}
\end{table}

\subsection{Independent adjudication}
To cover the dimensions the deterministic study cannot---above all the English
gloss---and to obtain a holistic verdict, we drew a seeded, stratified random
sample of 60 bh\=a\d{s}ya surfaces per band (180 total) and had it judged by
independent adjudicators \emph{blind to the confidence band}, each rating the
split, grammar, and meaning of every item as \emph{acceptable}, \emph{minor
issue}, or \emph{wrong}. Table~\ref{tab:adj} reports the results (175 items
returned). Again the bands rank-order cleanly: high- and medium-confidence
analyses are essentially always at least acceptable, and \emph{every} verdict of
``wrong'' fell in the low-confidence band---precisely the set the review loop
routes to a human. We report this adjudication as a secondary, corroborating
signal only: the adjudicator is itself a language model, so its verdicts risk
errors correlated with the analyser's, and human expert review (\S\ref{sec:review})
remains the gold standard.

\begin{table}[htbp]
\centering
\small
\begin{tabular}{@{}l r r r@{}}
\toprule
\textbf{Band} (\emph{n}) & \textbf{Split} & \textbf{Grammar} & \textbf{Meaning}\\
\midrule
High (58)   & 100.0 (98.3) & 100.0 (100.0) & 100.0 (100.0)\\
Medium (60) & 100.0 (98.3) & 100.0 (95.0)  & 100.0 (98.3)\\
Low (57)    & 98.2 (87.7)  & 96.5 (80.7)   & 96.5 (87.7)\\
\midrule
All (175)   & 99.4 (94.9)  & 98.9 (92.0)   & 98.9 (95.4)\\
\bottomrule
\end{tabular}
\caption{Independent, band-blind adjudication of a stratified sample: percentage
\emph{acceptable} (rated ``acceptable'' or ``minor issue''), with the strict
\emph{fully-correct} percentage in parentheses. Every ``wrong'' verdict occurred
in the low-confidence band.}
\label{tab:adj}
\end{table}

\subsection{Coverage and provenance of the evaluation}
Both evaluations are honest about their reach. The deterministic morphology check
scores only the $\approx$27\,\% of single-pada nominals that appear in the
Heritage lexicon as whole forms, and the segmentation check only the multi-pada
surfaces attested in the reference corpus; compounds and rarer derivatives, which
are harder cases, are under-represented in the covered subset. The adjudication
sample is small (180) and machine-judged. Neither replaces systematic human
expert review, which is ongoing through the correction loop and whose accumulated
verdicts will form the gold standard for future, larger studies. What the two
evaluations \emph{do} establish, jointly and by independent routes, is that the
analyses agree with established Sanskrit resources at a high rate, that quality
degrades gracefully and predictably with the confidence label, and that the
low-confidence worklist captures the errors. Finally, the built readers were
smoke-tested by driving a headless browser over the full corpus to confirm that
word-click, split- and lemma-aware search, citation navigation, and highlighting
all function with no runtime errors.

\section{Discussion and Limitations}
\label{sec:limitations}

\paragraph{Context insensitivity of the commentary dictionary.}
Analysing each distinct commentary surface once (``Route B'') buys scale and
consistency but cannot disambiguate a homographic form by its local
context---a surface that is genuinely ambiguous receives a single, most-likely
analysis wherever it occurs. The root text, analysed per-occurrence in verse
context, does not have this limitation. For the commentary, context-sensitive
disambiguation of the ambiguous minority is a clear avenue for refinement, and
the review overlay can already carry hand-disambiguated readings.

\paragraph{The LLM as scaffold.}
The analysis leans on an LLM, which can err confidently. Our harness---
deterministic-first resolution, adversarial refutation, mechanical rejoin
checks, confidence grading, and expert review---is designed to contain this,
but does not eliminate it. The confidence labels are self-assessed and should
be read as triage signals, not calibrated probabilities; calibrating them
against expert verdicts is future work.

\paragraph{Lemma search as a research instrument, and its limits.}
The lemma index makes the edition usable as a \emph{concordance} of the
\Prasthana{} and of \Sankara{}'s commentary---retrieval by dictionary headword
across inflection, \iast{sandhi}, and compounding---which we regard as one of the
resource's most useful capabilities for scholarship, since it lets a term or
concept be traced through the entire corpus in the form a scholar actually looks
it up. Its recall is, however, bounded by the per-word analysis it is built on: a
form whose lemma was mis-assigned---more likely in the low-confidence tier
(cf.\ \S\ref{sec:eval})---will be missed or mis-filed by a lemma query, so search
completeness improves exactly as the review loop corrects the underlying analyses.

\paragraph{Reproducibility.}
Every artefact is built from a single source corpus by a documented sequence of
scripts, and the built files are never hand-edited. The deterministic resources
(the inflectional lexicon, the corpus \padaccheda{}, the lexica) are public.
The LLM stage is the one component whose output is not bit-for-bit
reproducible; the expert-review overlay, however, is a permanent, human-authored
record that is reapplied deterministically, so the scholarly value added by
review is fully reproducible and portable across regenerations.

\paragraph{Licensing and provenance.}
The resource incorporates third-party data---an inflectional lexicon derived
from the Heritage engine, glosses from the Cologne lexica, and \padaccheda{}
attestations from public corpora---each under its own terms. Any public release
must and will carry the corresponding attributions and licences; the inflected
lexicon in particular is redistributed under its source's copyleft terms.

\section{Future Work}
\label{sec:future}
\begin{itemize}[leftmargin=1.4em,itemsep=0.15em]
  \item \textbf{Scaling the accuracy study} from the intrinsic evaluation of
        \S\ref{sec:eval} to a large \emph{human} expert study with
        inter-annotator agreement, using the review loop as the sampling frame
        and extending coverage to the compounds and rare forms that the
        lexicon-based checks under-represent.
  \item \textbf{Context-sensitive disambiguation} of ambiguous commentary
        surfaces, moving the most-recurrent ambiguous forms from Route B to
        per-occurrence analysis.
  \item \textbf{Confidence calibration} of the self-assessed labels against
        accumulated expert verdicts.
  \item \textbf{Interoperable export} in a standard scholarly encoding (e.g.\
        TEI with linguistic annotation, or a CoNLL-U layer) so that the analysed
        corpus can feed other tools, not only this reader.
  \item \textbf{Broadening the review community}, so that corrections from many
        Sanskritists accumulate into a shared, citable, versioned overlay.
\end{itemize}

\section{Conclusion}
We have presented an open, offline, word-level digital reader of the entire
\Prasthana{} with \Sankara{}'s \bhasya{}, in which every word of both root text
and commentary resolves to its euphonic split, grammatical analysis, and gloss,
and in which---because every word is lemma-tagged---the entire corpus is
searchable as a concordance by dictionary headword, across inflection,
\iast{sandhi}, and compounding.
The resource is, to our knowledge, the first to provide a uniform, interrogable,
word-level apparatus across the running prose of \Sankara{}'s commentary at this
scale (95{,}587 distinct forms over thirteen commentarial units). Its method
pairs deterministic Sanskrit resources with LLM-assisted analysis under an
adversarial verification harness, and closes with a durable human-expert review
loop so that scholarly correction is cheap, permanent, and cumulative. Offered
as a single self-contained file, freely redistributable and requiring no
infrastructure, it is intended above all to widen access---to bring the
foundational texts of Advaita Ved\=anta within word-level reach of students and
readers who lack years of grammatical training, while giving scholars a
platform they can correct and make their own.

\section*{Availability}
A live instance of the reader is publicly available at \liveurl.
It is distributed as a single self-contained HTML file that runs in any modern
web browser with no server, installation, or network connection. The build
scripts, source corpus, and derived data are available from the author and are
intended for open release under terms compatible with the incorporated
third-party resources.

\section*{Acknowledgements}
This work builds on the freely shared labour of the wider Sanskrit
computational and philological community, including G\'erard Huet and the
Sanskrit Heritage project, the Digital Corpus of Sanskrit, and the Cologne
Digital Sanskrit Dictionaries project, without whose open resources it would
not have been possible. The author thanks the Ramakrishna Mission Vivekananda
Educational and Research Institute for its support.


\end{document}